\documentclass{article}
\usepackage{cite}
\usepackage[dvips]{graphicx}
\usepackage{amsmath,amssymb}

\usepackage{algpseudocode,algorithm,algorithmicx}
\algrenewcommand\algorithmicrequire{\textbf{Input:}}
\algrenewcommand\algorithmicensure{\textbf{Output:}}

\usepackage{array}
\newcolumntype{L}[1]{>{\raggedright\let\newline\\\arraybackslash\hspace{0pt}}m{#1}}
\newcolumntype{C}[1]{>{\centering\let\newline\\\arraybackslash\hspace{0pt}}m{#1}}
\newcolumntype{R}[1]{>{\raggedleft\let\newline\\\arraybackslash\hspace{0pt}}m{#1}}

\renewcommand{\vec}[1]{{\bf{#1}}}
\usepackage{color}

\begin{document}
\title{Liveness Detection Using Implicit 3D Features}
%
%
%

\author{J. Matias Di Martino, Qiang Qiu, Trishul Nagenalli and Guillermo 
Sapiro}%
%

\maketitle

\begin{abstract}
Spoofing attacks are a threat to modern face recognition systems. In this work we present a simple yet effective liveness detection approach to enhance 2D face recognition methods and make them robust against spoofing attacks. We show that the risk to spoofing attacks can be reduced through the use of an additional source of light, for example a flash. From a pair of input images taken under different illumination, we define discriminative features that implicitly contain facial three-dimensional information. Furthermore, we show that when multiple sources of light are considered, we are able to validate which one has been activated. This makes possible the design of a highly secure active-light authentication framework. Finally, further investigating the use of 3D features without 3D reconstruction, we introduce an approximated disparity-based implicit 3D feature obtained from an uncalibrated stereo-pair of cameras. Validation experiments show that the proposed methods produce state-of-the-art results in challenging scenarios with nearly no feature extraction latency.
\end{abstract}

\section{Introduction}
Face recognition is one of the most common and reliable 
biometric techniques. In particular, 2D face recognition algorithms (i.e., 
methods that require a single standard camera) have become very popular in 
recent years since they can be readily used on personal computers and 
smartphones. 2D face recognition has become an accurate and technologically 
practical biometric alternative due in part to the massive amounts of labeled 
training data that social media and smart devices made available. Despite this, 
spoofing 2D attacks proved to be an actual relatively easy to deploy threat to 
modern face recognition systems. Different hacking methods have been developed 
in recent years to achieve illegal access to locked computers or smartphones 
guarded with face recognition technologies. One of the most popular and simple 
hacking techniques consists of printing or reproducing by some means a high 
quality picture of the target subject. Single cameras can not distinguish 
between a high definition replication of a subject picture and the live subject 
itself. Geometrically, those two situations are mathematically equivalent from 
the point of view of a single camera \cite{hartley2003multiple}. As a result of 
this, 3D systems are starting to be proposed, these being more difficult to hack 
but at the same time requiring significantly more sophisticated and less common 
hardware, are computationally and power more expensive, and suffer from the 
significantly reduced availability of 3D training data.

This work presents effective and yet simple alternatives to enhance 2D face recognition methods and make them more robust against a variety of spoofing attacks. We show that spoofing attacks can be reduced through the use of an additional device, e.g., a source of light (or flash) or another camera (as it is becoming more and more common even in mobile devices). First we analyze the use of an additional source of light. Combining two images of a live subject (taken under different illumination conditions), we show that implicit three-dimensional features of the subject face can be extracted highly efficiently, without the expensive intermediate step of 3D reconstruction, as commonly found in other systems. Moreover, we show that this approach can be extended to use multiple light sources, and learn a particular response model for each one of them. This can be used to design increasingly reliable face (liveness) active validation environments. Under the same framework of exploiting 3D features without 3D reconstruction, we also consider using two cameras. While it is well known that three dimensional information of a scene can be extracted from a stereo pair of cameras, the calibration of such pair of cameras and the triangulation of multiple views in the three dimensional world are yet computationally too expensive for this particular practical application. Therefore we propose an approximated (uncalibrated) estimation of the perceived local disparity which allows us to extract the needed implicit 3D features of the face with sufficient accuracy for the task of liveness detection. The proposed framework of 3D liveness detection without explicit 3D reconstruction is shown to produce state-of-the-art results at virtually no computational cost.

\subsection{Related work}\label{sec:RelatedWork}
A wide and ingenious variety of methods have been proposed to prevent spoofing attacks of face recognition systems. Diverse techniques may have significantly different hardware and computational requirements. For example, some methods make use of visible and thermal infrared imagery \cite{socolinsky2003face}, while other techniques recognize spontaneous eye blinks using a single low cost RGB camera \cite{pan2007eyeblink}. 

Methods based on multi-spectral analysis are among the most reliable liveness detection methods. For example, Pavlidis and Symosek \cite{pavlidis2000imaging} proposed a dual-band fusion system in the near infrared ($1,4-2,4\,\mu m$), and showed that human skin can be accurately detected by measuring the reflected light components in these particular spectral bands. Similarly, Zhang et al. \cite{zhang2011face} analyzed multi-spectral properties of human skin versus non-skin. Using a specific database they learned the discriminative wavelengths and trained a support vector machine (SVM) algorithm to fit the multi-spectral distribution for a Genuine-or-Fake classification. Socolinsky et al. \cite{socolinsky2003face} present a survey of multiple methodologies based on visible and thermal infrared imagery. Their analysis showed that under many circumstances, multi-spectral methods outperform methods that use standard cameras, while in some conditions both strategies can give equivalent results. They also showed that the performance can be improved when algorithms on visible and thermal infrared imagery are combined. Clearly multi-spectral methods can not always be applied, and the need for specific hardware restricts their use in many (consumer-type) practical applications.

Another family of methods use multi-modal approaches to discriminate between authentic and impostor faces. For example, Frischholz and Dieckmann \cite{frischholz2000biold} proposed the use of dynamic face, voice, and lip movement information to design a robust feature representation. In a similar way, Chetty and Wagner \cite{chetty2006multi} proposed to use audio plus video records to discriminate between authentic and imposter identities. As with the multi-spectral methods, the accuracy of these techniques comes at the cost of needing additional hardware components plus user cooperation.    

A popular class of liveness detection methods include the analysis of short videos from which 3D information of the face can be extracted. These methods became popular in the past decade since no specific hardware is required and they could be readily used on popular devices such as cellphones or personal computers. An example is the method proposed by Kollreider et al. \cite{kollreider2005evaluating}, who analyzed the trajectories of different parts of the face to discriminate authentic live faces against spoofed ones. Similarly, Bao et al. \cite{bao2009liveness} learned the differences in the corresponding optical flow fields comparing the dynamics of a three-dimensional face against that of a moving two-dimensional photograph. A related technique was proposed by Wang et al. \cite{wang2013face}, who proposed to detect spoofing attacks by recovering the sparse 3D information of the facial structure. To that end, they analyzed a face video or several images captured from more than two viewpoints. The experiments presented by Wang et al. show that this method is capable of detecting planar photo attacks as well as warped photos of a face in a robust way. It is also, however, substantially more complex than the previously mentioned ones both in terms of the number of steps and parameters that need to be estimated. Pan et al. \cite{pan2007eyeblink} also presented a liveness detection solution based on the analysis of a video sequence. Instead of extracting 3D structural features, they recognized spontaneous eye-blinks to discriminate between a real face and a spoof attack. This simple idea, although not as robust as some of the previous solutions, allows having real-time detection schemes with relatively low computational and hardware complexity.

An additional set of works require a single input image and exploit different properties of this single image to detect the differences between photos of authentic live faces and those of spoofing attacks. For example, Li et al. \cite{li2004live} compared the Fourier spectra of authentic and fake images to differentiate live faces from impostor ones. They observed that (in some particular cases) the high frequency components of a printed picture are less intense than of a live face. A significantly more elaborated method was proposed by Tan et al. \cite{tan2010face}, who modeled faces as Lambertian surfaces and applied this model to infer structural information from a single input image. Inspired by the ideas behind illumination invariant face recognition, they used different prior models to decompose the input face into an albedo and an illumination component. Then, they inferred the 3D structural information of the face from the illumination component, since it contains information about the surface normals. In addition, they used a Difference of Gaussian method based on the intuitive idea that the image of a photograph taken through a webcam is essentially an image of a real face that passes through the camera system twice and the printing system once. As Li et al., they suggested that printed images will present lower high-frequency components compared to the images of a live face. Peixoto et al. \cite{peixoto2011face} extended the work of Tan et al. for scenes in which the illumination conditions are poor. To that end, they pre-process the input images using the contrast-limited adaptive histogram equalization algorithm \cite{zuiderveld1994contrast} and then replicate similar steps to the ones proposed by Tan et al. A different solution was proposed by Yeh and Chang \cite{yeh2017face}. Their method relies on the optical properties of lenses when they are focused on a particular plane with a narrow depth of field. Two features, the blurriness level and the gradient magnitude threshold, were computed on the nose and the cheek regions of the face. The differences of these two features across different regions of the face are then used for classification. Specifically, they proposed to acquire the input image by focusing the camera on the nose, since for a real face the nose will be sharp but other regions such as the cheek will be blurred. Although single image methods may be effective for some particular applications, they can be easily hacked as they make strong assumptions about the observed scene. For example, the method proposed by Yeh and Chang can be hacked by simply taking a photo in the same testing conditions (a narrow depth of field) and then printing it (a simple post-processing of an image can achieve the same effect). This would create a photo where only the nose is perceived as sharp, even though we are displaying in front of the camera an impostor (planar) image of a face.

Recently, Chan et al. \cite{chan2018face} proposed the use of two images, one taken under ambient illumination $I_a$ and the other one using a flash light $I_f$. In their work, they suggested that this approach can enhance the differentiation between legitimate and illegitimate users while reducing the influence of environmental factors. The face region and the background are detected in both input images $I_a$ and $I_f$, and these two input images are then analyzed with four texture descriptors. The first proposed descriptor consist of a Local Binary Pattern (LBP) \cite{wang1990LBP} of the face patch in image $I_f$ (picture taken with the flash on). This descriptor is applied across nine regions of the face, leading to a $531$ dimensional feature vector. The second descriptor is the standard deviation of the difference between the face patches on $I_a$ and $I_f$ respectively. The third descriptor corresponds to the mean value of the difference between the background patches on $I_a$ and $I_f$. The final descriptor is defined as the standard deviation of the difference between background patches on $I_a$ and $I_f$ respectively. Our work is closest to this technique and we will provide more details and a full comparison with it in the experimental section.

In the present work we exploit the idea of efficiently computing 3D features without explicit reconstruction. We first propose a solution based on analyzing two images taken under different lighting conditions. For example, one can take the first image under ambient illumination and a second image using a flash light as proposed by Chang et al. (although our analysis is not restricted to that particular case). Although the structure of the input data that we consider is similar to the one considered by Chang et al., we exploit these input images in a substantially different way. In Section \ref{sec:Experiments} we highlight these differences, and we provide a complementary discussion of the main strength and weakness of the different methods. We then extend this technique to the case of multiple light sources and propose a significantly more secure liveness detector based on active illumination. To conclude on the idea of 3D features without 3D reconstruction, we consider a very low cost stereo approach. We demonstrate that the proposed approach produces state-of-the-art results at virtually no additional computational cost. We also show that new subjects no present in the training data are efficiently handled, and so are different scenarios not used at all for training. 

\section{Implicit 3D Feature Extraction}
\begin{figure*}
\centering\includegraphics[width = \textwidth]{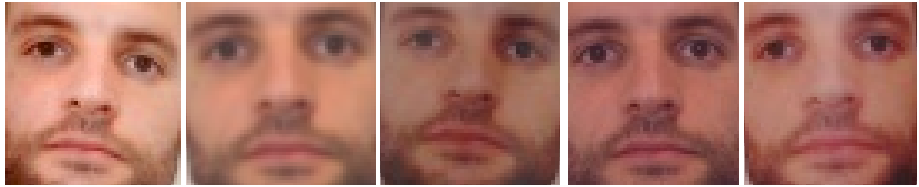}
\caption{From left to right, the second and fourth pictures correspond to a photo of the actual (live) subject. The first picture is a photo of a computer screen where the face of the test subject is displayed. The third picture was taken by photographing a printed portrait of the test subject, and finally the fifth picture is a photo of the same print but now bended.}\label{fig:illustrativeFaces}
\end{figure*}

Could the reader identify which one of the pictures in Fig.~\ref{fig:illustrativeFaces} corresponds to a photo of the real (live) subject? Only pictures two and four (from left to right) are photos of the actual person, the other ones where taken of a print or screen. Detecting spoofing attacks from a single image is a very hard problem and requires strong assumptions on the scene (e.g., that the camera has a finer resolution than the display/printer \cite{li2004live}, or that the texture of the facial image fits a particular face model \cite{tan2010face}). In this work we address this liveness detection challenge with a new framework that exploits more than one image to implicitly introduce 3D information. Contrary to the usual literature, these multi-images are obtained with standard and ubiquitous devices.  

\subsection{Illumination based features} 
We now show how 3D implicit features can be extracted from a pair of images taken under different lighting conditions, e.g., flash/no-flash, without explicit 3D reconstruction.

Let us start by describing how light interacts with different surfaces (a face in our case). This will allow us to subsequently engineer simple features that implicitly capture the 3D information of a given surface. In addition, we show how multiple images taken under different illumination conditions can be used to infer 3D information without actually retrieving the 3D shape of the scene. 

\begin{figure*}
\centering \includegraphics[width = .75\textwidth]{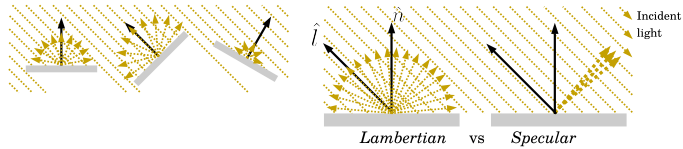}
\caption{Illustration of a \emph{Lambertian} and a \emph{Specular} surface. On the left we illustrate how, for a Lambertian surface, the intensity of the light reflected by the surface depends on the orientation of the surface with respect to the direction of the incident light. On the right, we illustrate the difference between a Lambertian reflection and a Specular reflection.}
\label{fig:lambertianSurface}
\end{figure*}

Surfaces reflect light differently depending on their microscopical properties. The two more popular models for describing surface reflection are the \emph{Lambertian} and the \emph{Specular} ones. A Lambertian surface reflects light by scattering it in all possible directions as illustrated in Fig.~\ref{fig:lambertianSurface}. In contrast, when a ray of light reflects over a specular surface, the reflected light has the same angle as the incident light. An example of a reflecting surface is a mirror. Most natural surfaces can be modeled as a combination of these two ideal models, where one might refer to the \emph{Lambertian} or \emph{Specular} component of a given surface. 

In the present work, we focus on the study of liveness detection by exploiting faces' reflection properties. Faces can be modeled as Lambertian surfaces \cite{Zhou2007,hayakawa1994,shashua1997,basri2003lambertian}. Since Lambertian surfaces scatter light in all possible directions, the intensity of the perceived light remains invariant when we move our eyes (or camera) in front of the object. However, the relative orientation of the surface with respect to the direction of the incident light does play a significant role. It is easy to show that the intensity of the reflected light is proportional to the cosine of the angle between the incident light and the surface normal, as we illustrate in Fig.~\ref{fig:lambertianSurface}. This can be expressed as
\begin{equation}\label{eq:reflectancePoint}
I = I_0\,a\, (\vec{\hat{n}} \cdot \vec{\hat{l}}),
\end{equation}
where $I_0$ corresponds to the intensity of the incident light, $a$ represents the surface albedo,\footnote{This is defined as the ratio of the irradiance reflected and the irradiance received by a surface.} $\vec{\hat{n}}$ is a unit vector normal to the surface, and $\vec{\hat{l}}$ a unit vector indicating the direction in which the light rays approach the surface.

Complex surfaces like faces have different local properties. Hence Eq.~\eqref{eq:reflectancePoint} must be formulated for each particular point $(x,y,z)$ of the face,
\begin{equation}\label{eq:reflectanceFace1}
I_{(x,y,z)} = I_0\,a_{(x,y,z)}\, (\vec{\hat{n}}_{(x,y,z)} \cdot \vec{\hat{l}}),
\end{equation}
where we assume an homogeneous illumination ($I_0,\ \vec{\hat{l}}$). For compactness, we include the intensity factor $I_0$ into the vector $\vec{\hat{l}}$ (i.e., $\vec{l} \stackrel{def}{=}I_0\vec{\hat{l}}$). 

So far we considered the case of a single source of (planar) light illuminating the surface. The extension to multiple light sources $\vec{l_1},\ \cdots,\ \vec{l_m}$ is straightforward,
\begin{equation}\label{eq:reflectanceFace2}
I_{(x,y,z)} = a_{(x,y,z)}\,\sum_{i=1}^m \max\left(\vec{\hat{n}}_{(x,y,z)} \cdot \vec{l_i},\ 0\right).
\end{equation}
The $\max()$ operation in Eq.~\eqref{eq:reflectanceFace2} is introduced to impose that only the light sources that are \emph{in front} of the surface can actually illuminate it. A special case is of particular interest; if an object is homogeneously illuminated from all possible directions $(\theta,\varphi)$, Eq.~\eqref{eq:reflectanceFace2} becomes
\begin{align}\label{eq:reflectanceH}
I_{(x,y,z)} & = a_{(x,y,z)}\,\int\int \max\left(\vec{\hat{n}}_{(x,y,z)} \cdot \vec{l}_{(\theta,\varphi)}\,,\ 0\right)d\theta d\varphi, \\
 & \propto a_{(x,y,z)}. 
\end{align}
It is easy to prove that Eq.~\eqref{eq:reflectanceH} is proportional to the surface albedo for all $\vec{\hat{n}}$. In other words, when a surface is illuminated by scattered light, the light reflected by the surface becomes independent of the 3D structure of the object and only the albedo properties are manifested.

\begin{figure*}
\centering\includegraphics[width = .9\textwidth]{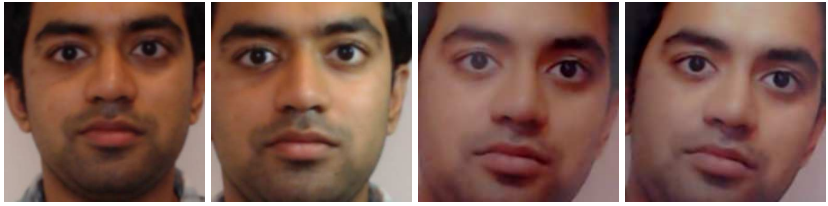}
\caption{From left to right: Picture of a face under ambient illumination (denoted as $I_a$), picture taken with an additional source light (denoted as $I_f$), and similarly two pictures under the same conditions where a printed photo replaces the live subject. The differences between these two pairs will be exploited to detect liveness.}\label{fig:FacesLight}
\end{figure*}

Figure~\ref{fig:FacesLight} illustrates a pair of images from a live subject (first two images, left to right) and a pair of images captured of a simulated spoofing attack. In both cases the illumination conditions changed between the shots.
The image taken under ambient illumination is denoted as $I_a$, and the image when an additional source of light is included as $I_f$. Assuming that the camera parameters (shutter speed, ISO, and aperture) are fixed between the shots, we can relate both pictures using the Lambertian face model previously introduced:
\begin{align}
I_a(u,v) = &\ a(u,v) \sum_{i=1}^m \max\left(\vec{\hat{n}}(u,v) \cdot \vec{l_i},\ 0\right), \label{eq:Ia} \\
I_f(u,v) = &\ a(u,v) \sum_{i=1}^m \max\left(\vec{\hat{n}}(u,v) \cdot \vec{l_i},0\right) \nonumber \\
           & + a(u,v)\ \vec{\hat{n}}(u,v) \cdot \vec{l_f}. \label{eq:If} 
\end{align}
Equations~\eqref{eq:Ia} and \eqref{eq:If} model the ambient illumination as an arbitrary (unknown) combination of $m$ light sources. The vector $\vec{l_f}$ in Eq.~\eqref{eq:If} represents the intensity and direction of the additional light source present only in the second shot. One may notice that we replaced the three-dimensional world coordinates $(x,y,z)$ used in eqs.~\eqref{eq:reflectanceFace1}-\eqref{eq:reflectanceFace2} by two dimensional pixel coordinates $(u,v)$. $I(u,v)$ represents the light collected by the pixel $(u,v)$ which corresponds to the optical image of the emitted light $I(x,y,z)$. 

We now define the quantity
\begin{equation} \label{eq:definitonOfQ}
I3D(u,v) \stackrel{def}{=} \frac{I_f(u,v) - I_a(u,v)}{I_a(u,v)}, 
\end{equation}
which is easy to prove is equivalent to
\begin{equation} \label{eq:I3D2}
I3D(u,v) = \frac{\vec{\hat{n}}(u,v)\cdot \vec{l_f}}{\sum_{i=1}^m \max\left(\vec{\hat{n}}(u,v) \cdot \vec{l_i} \right) }.
\end{equation}

\begin{figure*}
\centering\includegraphics[width = \textwidth]{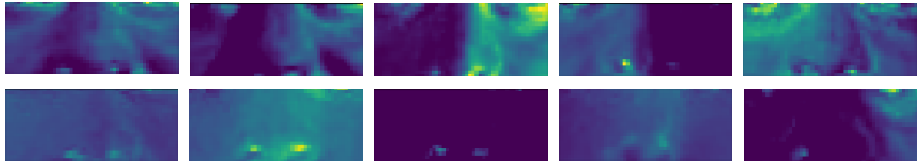} 
\caption{The quantity $I3D$ computed for a portion of the face for different test subjects and lighting conditions. The first row shows the value of $I3D$ on pictures of live subjects and the second row shows the exact same procedure for different spoofing attacks (printed or screen displayed). These clear visual differences in $I3D$ will be used to detect such attacks.} \label{fig:I3D}
\end{figure*}

In order to compute the lighting response $I3D$, a pair of registered input images $I_a$ and $I_f$ are required. The first step consists of detecting the face and facial landmarks in both images. To that end, the dlib \cite{king2009dlib} library is used. Once the landmarks are detected, the pair of images taken under different illuminations are registered through an homography. Then the proposed features are computed, according to Eq.~\eqref{eq:definitonOfQ}, over a selected predefined patch of the face, e.g., based on the same computed landmarks. Algorithm~\ref{alg:I3D_features} summarizes the main steps followed to compute the described feature vector. 
\begin{algorithm}
  \caption{Compute $I3D$ features}
    \label{alg:I3D_features}
  \begin{algorithmic}[1]
    \Require{Input images $I_{a}$ and $I_{f}$}
    \State Detect face and facial landmarks in both images.
    \State Set im. resolution (down sample to a fixed image size).
	\State Register $I_f$ over $I_a$ through an homography  
    \State Select a face patches $I_f^{patch}$, $I_a^{patch}$ 
	\State Compute $I3D = \frac{I_f^{patch} - I_a^{patch}}{I_a^{patch}}$ 
    \State \Return{$I3D$} 
  \end{algorithmic}
\end{algorithm}

$I3D$ has an interesting intrinsic property for the particular task at hand (in addition to its easy computation); it is independent of the surface albedo, making it less sensitive across different subjects, while coding 3D information about the shape of the surface. While it is not easy to retrieve the 3D shape of the surface from $I3D$ (which is not the goal, we are after liveness and not recognition), it is possible to use the information coded in $I3D$ as an implicit 3D features of the surface we are looking at. Figure~\ref{fig:I3D} illustrates $I3D$ for different subjects. The first row was obtained from a pair of images of live subjects, while the second row corresponds to pairs of images of different kinds of spoofing attacks. Both shape ($\vec{\hat{n}}(u,v)$) and illumination information are clearly coded in $I3D$.  

\subsection{Approximated disparity-based features} 
One of the most popular methods for capturing three dimensional information of an arbitrary scene consist of using a pair of (stereo) cameras. The relative shift between a stereo pair of images is related to the distance of the objects to the cameras \cite{hartley2003multiple}. While retrieving the absolute 3D information of the face provides very powerful features for the analysis of liveness, it requires the pair of cameras to be calibrated, which is time demanding. Following the idea presented above of enhancing liveness detection by inferring 3D information without explicit 3D reconstruction, we now propose an alternative to extract three dimensional cues from an uncalibrated pair of images. We assume that two images $\{I_{left}, I_{right}\}$ are obtained by a stereo pair of cameras (or by a single camera that moved between the frames). Left and right images can be related by
\begin{equation}\label{eq:approxDisparity1}
I_{left}(u,v) = I_{right}((u,v)+\vec{d}(u,v)),
\end{equation}
where the disparity vector $\vec{d}(u,v)$ is related to the depth $z(u,v)$. If calibration were to be performed and $\{I_{left},I_{right}\}$ are replaced by their rectified \cite{fusiello2008quasi,monasse2011quasi} counterparts, the disparity $\vec{d}$ can be computed for each pixel position $(u,v)$ by following a one dimensional patch matching procedure. As an alternative to this calibrated approach, we will estimate an approximated disparity, and demonstrate that this is sufficient for liveness detection. 

First the location of the tip of the nose (as detected with the same landmarks technique previously mentioned) is used to align both images through a global translation. Then we estimate the relative disparity by expanding the right term of Eq. \eqref{eq:approxDisparity1},
\begin{align}
I_{left}(u,v) & = I_{right}((u,v)+\vec{d}(u,v)) \nonumber \\
              & \approx I_{right}(u,v) + \nabla I_{right}(u,v)\cdot\vec{d}(u,v). \label{eq:approxDisparity2}
\end{align}
Assuming that both cameras were located approximately along the u-axis of the image sensors leads to
\begin{equation}\label{eq:approxDisparity3}
Da3D(u,v) \approx \frac{I_{left}(u,v)-I_{right}(u,v)}{\frac{\partial I_{right}(u,v)}{\partial u}}.
\end{equation}

The approximated disparity $Da3D$ descriptor can be computed from an uncalibrated pair of stereo images following very simple steps. First, the face in both images and the facial landmarks are detected. To this end, as before the dlib \cite{king2009dlib} library was used. Once the landmarks are detected, the stereo images are registered together through a rigid translation (making the landmarks associated to the tip of the nose to match). This rigid transformation allows us to preserve the relative disparity between different portions of the face. Then, the proposed features are computed over a predefined patch of the face,  extracted based on the already computed landmarks, following Eq.~\eqref{eq:approxDisparity3}. Algorithm~\ref{alg:approx_features} summarizes the main steps described above.
\begin{algorithm}
  \caption{Compute approximate disparity features}
    \label{alg:approx_features}
  \begin{algorithmic}[1]
    \Require{Input images $I_{left}$ and $I_{right}$}
    \State Detect face and facial landmarks in both images.
    \State Set im. resolution (down sample to a fixed image size).
	\State Translate $I_{right}$ to match the tip of the nose of $I_{left}$.
    \State Select a face patch $\{I_{left},I_{right}\}\leftarrow \{I_{left}^{patch},I_{right}^{patch}\}$.
   	\State Compute $u$-partial derivative $I_{right}^{u}$.
    \State Compute $Da3D = \frac{I_{left}-I_{right}}{I_{right}^{u}}$ (if $I_{right}^{u}\approx 0$ set $Da3D=0$).
    \State \Return{$Da3D$.}
  \end{algorithmic}
\end{algorithm}

\section{Experimental Results}\label{sec:Experiments}
Due to the lack of publicly available databases with the lighting conditions needed to perform liveness detection, we created test data (to be released upon acceptance) by photographing 17 subjects in different environments and lighting conditions. Since the goal of this work is to distinguish a real face from different spoofing attacks, rather than differentiate between people as is the case of face recognition, as here demonstrated the database size requirement is less demanding compared to the richness of variations in the illumination conditions and acquisition conditions \cite{tan2010face}. We collected $5,010$ pairs of stereo images plus $7,503$ pairs of images under different lighting conditions. A pair of webcams \emph{Logitech C920 HD} were used to collect images of real (live) subjects, printed pictures, and screen-displayed portraits, simulating different spoofing attacks. Approximately half of the samples correspond to live faces and the other half to various types of attacks. In addition, for printed faces many different deformations of the shape of sheets were used during the data collection. The environments in which the images were collected presented significant difference in the illumination conditions. The ambient luminance varied from 100 to 400 Lux and the additional source of light was set to increase the luminance over subjects faces between 20-50\%. Additional images were collected, as it will be further detailed below, with mobile phones for testing the system's robustness to conditions not in the training data.

\subsection{Analysis of different features and classification methods}
Table~\ref{tb:features_and_classifiers} summarizes the classification (live vs. attack) accuracy for different sets of features and classification methods. For classification, we tested both Support Vector Machine classifiers and a light-weight convolutional neural network (CNN), its architecture is provided in Table~\ref{tab:network-arch}. The experiments were performed using the entire database described before, selecting 80\% of the image pairs for training and the remaining ones for testing. This process was repeated 10 times, and the mean and standard deviation test accuracy is reported. The pair of stereo cameras was calibrated and the result using the disparity computed from the pair or rectified images is also included for comparison (remember that such calibration is not performed in our proposed system).   
\begin{table}
\begin{center}
\caption{Mean accuracy and standard deviation for different sets of features and classification methods. The experiments were repeated ten times choosing random subsets for training ($80\%$) and testing ($20\%$). For comparison, the first row presents the results with full stereo calibration. The evaluation time includes the time for the computation of facial landmarks, the features, and the classification of an individual sample. The fourth column (Extraction time) accounts for the feature extraction time only.}
\label{tb:features_and_classifiers}
\begin{tabular}{L{3.2cm} C{1.5cm} C{.8cm} C{.8cm}}
\hline\noalign{\smallskip}
  Method Description & Acc. (\%) & \ Eval. time (ms) & Ext. time (ms)\\
\hline\noalign{\smallskip}
 SVM + Calibration-Face & $96.7\pm3.9$ & $2,228$ & $2,228$\\
 \hline \noalign{\smallskip}
 SVM + Chan et al. \cite{chan2018face} & $96.8\pm0.5$ & $54$ & $29$\\
 \hline \noalign{\smallskip}
 SVM + $I3D$-Face & $99.4 \pm 0.3$ & $29$ & $4$ \\
 SVM + $I3D$-Nose & $96.9 \pm 0.7$ & $29$ & $4$ \\
 SVM + $Da3D$-Face & $95.7 \pm 1.1$ & ${\bf{25}}$ & $\bf{0}$ \\
 SVM + $Da3D$-Nose & $78.9 \pm 1.1$ & ${\bf{25}}$ & $\bf{0}$ \\
 \hline \noalign{\smallskip}
 CNN + $I3D$-Face & ${\bf{99.8 \pm 0.0}}$ & $29$ & $4$\\
 CNN + $Da3D$-Face &  $97.8 \pm 0.0$ & $\bf{25}$ & $\bf{0}$ \\
\noalign{\smallskip}
\hline
\end{tabular}
\end{center}
\end{table}

As part of the experimental study, we compare with the closest work to ours \cite{chan2018face}. The features proposed by Chan et al. can be summarized (as mentioned in the introduction) as: (i) an LBP descriptor of the image taken with flash, (ii) the standard deviation over the region of the face of the difference between the image taken with and without flash, and (iii) features extracted from the background of the scene. Since we are interested in solutions that could be applied in arbitrary scenes where the background of the photos can not be controlled (e.g., mobile devices), we considered face only features (ignoring the background of the images). Results are also reported in Table~\ref{tb:features_and_classifiers}.

\begin{table}
\begin{center}
\caption{\label{tab:network-arch}
Light-weight convolutional neural network (CNN) used in our liveness detection experiments. It takes around $0.3ms$ on a (desktop) CPU to evaluate a $28 \times 28$ implicit 3D feature. FC stands for fully-connected. }
\begin{tabular}[t]{ l}
\hline
conv $5\time 5\times 1\times 16$, ReLu, max-pooling $3\times 3$\\
conv $5\times 5\times 16\times 64$,  ReLu,  max-pooling $3\times 3$ \\
  FC $128$,  ReLu,  FC $2$ \\
\hline       	
\end{tabular}
\end{center}
\end{table}

We tested the computational time using a CPU Intel Core i5 (with the exception of the calibrated disparity method that was tested on a Intel Core i7 CPU processor). The detection of the facial landmarks required about $25ms$, and the featured proposed by Chan et al. took an extra $29ms$ (total time $54ms$). Computing $I3D$ required almost an order of magnitude less, $4ms$ after the landmarks detection (total time $29ms$), and $Da3D$ takes a negligible amount of time (total time $25ms$). For the sake of comparison, we implemented a standard stereo pipeline, i.e., for a pair of stereo images the epipolar rectification was performed (using calibration data previously obtained), and the disparity was then computed through a patch based local image matching. The total time required per pair of images to compute the facial disparity map was about $2,228ms$, this is about $80$ times the time required for the proposed technique, illustrating the value of our approach, which obtained basically the same performance (Table~\ref{tb:features_and_classifiers}).

\begin{figure*}
\centering\includegraphics[width = \textwidth]{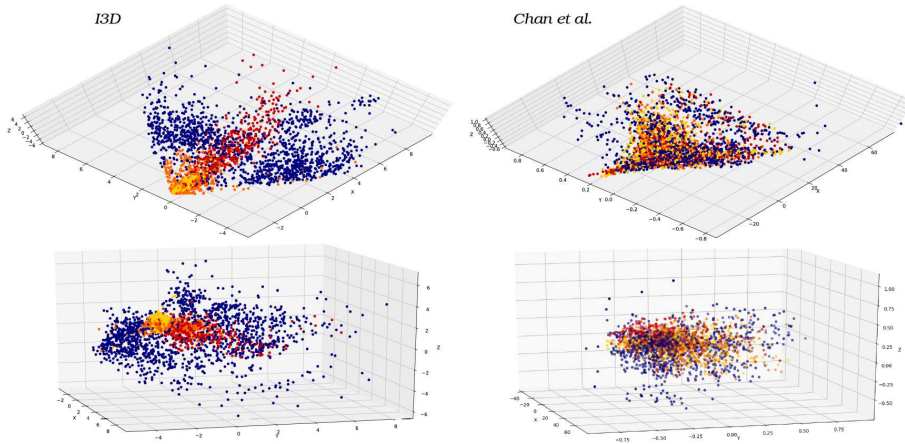}
\caption{PCA three dimensional embedding of $I3D$ features (left column) and the facial features proposed by Chan et al.  \cite{chan2018face} (right column). Each row corresponds to a different view of the same three-dimensional space. Blue dots represent samples of pair of pictures taken from real live faces, the red, orange and yellow dots represent spoofing attacks by using a flat print, a curved print, and screen-displayed face respectively.}\label{fig:PCA}
\end{figure*}

Complementing the previous analysis, Fig.~\ref{fig:PCA} illustrates a three dimensional PCA embedding of the proposed sets of features. 

To recap these results, the proposed framework achieves state-of-the-art performance at (often significantly) reduced computational cost. We will  demonstrate next that the advantages of this framework manifest themselves also in robustness.

\subsection{Evaluation of the training data properties}
\subsubsection{Impact of the number of samples}
\begin{figure}
\centering\includegraphics[width = \columnwidth]{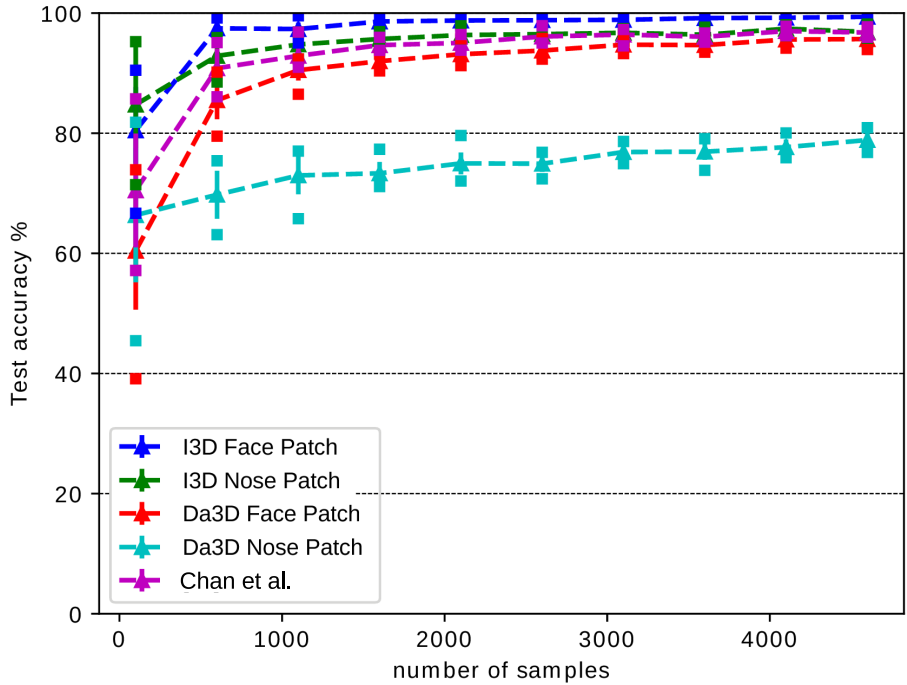} 
\caption{Performance of the SVM classifier for different sets of features when a different number of samples are used for training. In all the cases the accuracy over the test set is reported. The bars correspond to the standard deviation obtained over ten random repetitions of the experiment. The mean value over the random set repetitions is plotted using a triangular marker, while the minimum, and the maximum values are plotted as square markers.} \label{fig:SVM}
\end{figure}

Fig.~\ref{fig:SVM} presents the accuracy over the test set for five different sets of features. In this experiment the SVM algorithm was used for classification. The blue and green curves were obtained using $I3D$, first using a larger patch over the face (as illustrated in Fig.~\ref{fig:I3D}), and then considering only the region of the nose. Red and cyan curves correspond to the approximate disparity features $Da3D$, again the same two regions of the face were tested. Finally, the violet curve shows the results obtained with the facial features proposed by Chan et al. As we can see, a few thousand samples are enough to achieve an accuracy over 90\%. The best results were achieved for the illumination based $I3D$ features extracted over the face patch. The region of the nose has very rich shape information when the $I3D$ features are considered. On the other hand, $Da3D$ features require a larger portion of the face for a successful discrimination. The pair of stereo cameras we tested were located relatively close to each other (the distance between the camera sensors was about $4cm$). The baseline between the camera sensors plays an important role in the depth resolution \cite{hartley2003multiple} and may explain why the approximate disparity features perform better for our specific setup when they are computed for a larger area of the face. We would like to study in the future even closer cameras, as those are appearing today on mobile phones.

\subsubsection{Data resolution}
Table~\ref{tb:resultsLR} shows the accuracy (ACC) and Half Total Error Rate (HTER) \cite{chan2018face} using $I3D$ as feature versus the two types of facial features proposed by Chan et al. This experiment was performed 10 times randomly sub-sampling 2k pairs of images from which 80\% was used for training and the remaining 20\% for testing. An SVM was used for classification.  
\begin{table}
\begin{center}
\caption{Accuracy and HTER (\% values) for different sets of features and input image resolutions. The proposed features are robust under different resolutions, while texture based features are more sensitive to resolution variations.}
\label{tb:resultsLR}
\begin{tabular}{L{1.4cm} C{1.4cm} C{1.3cm} C{1.4cm} C{1.3cm}}
\hline\noalign{\smallskip}
Input res.:  & \multicolumn{2}{c}{$(1080\times 1920)$} & \multicolumn{2}{c}{$(270\times 480)$} \\
  & Acc & HTER & Acc & HTER \\
\hline\noalign{\smallskip}
\cite{chan2018face} & $97.9\pm0.9$ & $2.1\pm0.9$ & $94.9\pm0.8$ & $5.1\pm0.8$ \\
$I3D$-Face & $\bf{99.5\pm0.2}$ & $\bf{0.6\pm0.3}$ & $\bf{99.4\pm0.3}$ & $\bf{0.5\pm0.4}$ \\
\noalign{\smallskip}
\hline
\end{tabular}
\end{center}
\end{table}

This experiment is very important to illustrate one of the key differences between the features proposed here and the ones introduced by Chan et al. When high resolution input images are used, we retrieve similar results to the ones reported in \cite{chan2018face} (at a reduced computational cost). However, if we repeat the exact same experiment using a low resolution version of the original images, the performance of the features proposed in \cite{chan2018face} deteriorate more significantly than ours. This can be explained in a very intuitive way: 531 out of the 532 dimensions of the proposed feature vector correspond to the LBP texture descriptor obtained from the flash-image. Therefore, a significant portion of the feature space is capturing texture information, e.g., how sharp the real versus the fake images are. In contrast, the features proposed in this work focus on implicitly exploiting the three dimensional shape of live faces, which does not depend as strongly on the particular resolution the real and fake images might have. This robustness is critical to guarantee performance across sensing devices. In all the other experiments presented in this section, we used the low resolution version of the captured images to make each individual fake and real sample harder to distinguish from each other. 

\subsubsection{Novel subject and generalization capability}
In order to evaluate how easily the proposed approach can be generalized to new subjects (that were not present during the training process), the proposed $I3D$ features were evaluated using the standard leave-one-out procedure. As a classifier, we use the light-weight CNN model. The mean accuracy obtained was $97.8\%$, showing that the proposed approach can be applied on novel subjects. Note that in order to match the CNN input layer dimensions, the $I3D$ feature vector was first resized to a $28 \times 28$ CNN input, and it took the CNN $0.3ms$ on a desktop CPU to classify a feature sample.

\subsection{Differentiating light sources: Active Liveness Detection}
In the following set of experiments, we tested how different light models could be obtained from the input features $I3D$. We isolated from the entire database those samples for which the additional source of light came from the right/left (with respect to the test subject face). Using this subset of $623$ real samples (that contains 15 different subjects), we trained a new SVM model to distinguish between the pair of pictures taken under Ambient/Ambient+RightLight from those taken under Ambient/Ambient+LeftLight. Using $I3D$ features extracted from the region of the face and the nose we obtained an accuracy of $99.5\pm0.4$ and $99.4\pm0.5$, respectively. This is a very interesting result from a practical perspective, as it shows that the proposed features can be used to design reliable \emph{active} liveness validation environments. For example, by installing two flashes on the right and left of the camera respectively, one can actively and randomly select which source of light is being activated and in which order. As an example, we could randomly choose the sequence of activation \emph{Left-Left-Right-Left} and then validate if the captured sequences of images match the proper left-right models. This makes the liveness detector increasingly more robust, and the number of light sources can be adapted to the desired security level.

\subsection{Liveness detection on a smartphone}
\begin{figure*}
\centering\includegraphics[width = .9 \textwidth]{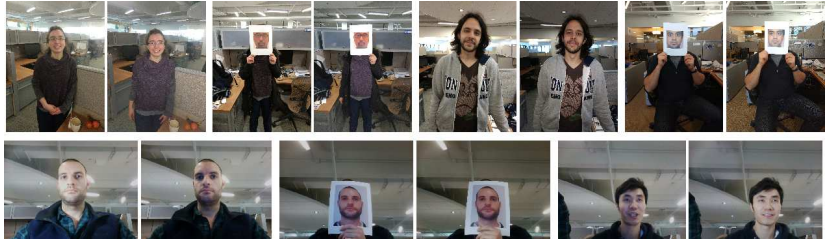}
\caption{The first row illustrates pairs of images (with and without flash) taken using a smartphone. The second row corresponds to pairs of images taken using a computer built-in webcam (and switching the desk light on and off).}\label{fig:examples}
\end{figure*}

A final set of experiments was carried out testing the proposed method on different devices. As described at the beginning of this section, the training data was collected using a commercial webcam model, the \emph{Logitech C920 HD}. This entire dataset is used to train a SVM model. To test the impact of using different acquisition devices without the need for retraining (meaning, system already deployed),  we collected new test data using the \emph{Moto G4} smartphone and the webcam in a \emph{Lenovo T430u} laptop. $52$ pairs of images (with and without flash) were collected using the phone camera, and 35 pairs of images were collected using the laptop webcam. Since the computer does not have a built-in flash, we changed the illumination conditions by toggling on and off a standard desk lamp. Figure~\ref{fig:examples} shows a subset of these collected new test images. It is important to highlight again that no training is performed using this new data. The classification model was fitted using the data previously collected (using different hardware, locations and illumination conditions). We computed the $I3D$ feature for each new pair of images and achieved a classification accuracy of $98.9\%$, so we observe that there is basically no deterioration in performance for this radically different test set. The $58$ spoofing attacks simulated were successfully detected and only one live face out of the $29$ tested was misclassified. This result shows robustness of the proposed framework and the potential of the proposed features to be used with ubiquitous devices and settings, even when no training data for a particular device is available. 

\subsection{Qualitative overview and discussion}
To conclude this section we provide a qualitative overview of some of the methods described in Section~\ref{sec:RelatedWork}. We focus on differences in the hardware requirements and robustness against different spoof attacks of each method analyzed. It is important to highlight that our analysis does not try to establish whether a method can be hacked or not, but how much work it takes to do so. 

Table \ref{tab:MethodsComparision} summarizes some of the key features of different liveness detection approaches. For example, in Section~\ref{sec:RelatedWork} we discussed how the technique proposed by Yea and Chang \cite{yeh2017face} (based on an out-of-focus analysis) can be hacked by taking a photo of a live face using a narrow depth of field and then printing it. Another interesting example of a potential spoofing attack can be elaborated by analyzing approaches that capture the 3D structural information of the face from its motion (e.g., \cite{wang2013face,kollreider2005evaluating}). When a tablet is used to display the face of a target subject, if the face is shown in fixed pose while the tablet is rotated, the motion of the facial landmarks will indicate that we are looking at a flat surface. In this situation, the attack can be successfully prevented. However, if the tablet remains static and we simulate the rotation of the subject's face using, for example, a generic 3D model, then the movement of the landmarks perceived by the camera would be indistinguishable from those produced by a \emph{3D face}. Techniques based on a monocular view can be improved by using an additional component. In this work we presented two such alternatives using either an additional camera or an active source of light. While the presented method is robust under a large variety of common spoofing attacks, it could be hacked by creating an accurate 3D replica of a subject face, which is of course more sophisticated than tools used in common attacks. To prevent this type of attack one can combine the ideas presented in this work with a facial video analysis of the subject. For example, one could design a system that measures the proposed set of features in real time while asking the test subject to perform different facial expressions which are then automatically detected and matched.

\setlength{\tabcolsep}{2pt}
\begin{table*}
\begin{center}
\caption{Properties of a set of liveness detection methods.}
\label{tab:MethodsComparision}
\begin{tabular}{L{.7cm} L{3cm} C{1.2cm} C{1.2cm} C{1.2cm} C{1.2cm}}
\hline\noalign{\smallskip}
  & Input Requirements  & \multicolumn{4}{c}{Could potentially be hacked by a:} \\
  &  & Planar Photo & Curved Photo & Tablet & 3D replica\\
\noalign{\smallskip}
\hline
\noalign{\smallskip}
\multicolumn{6}{c}{Multi-spectral} \\
\cite{socolinsky2003face} & Visible and thermal cameras & No & No & No & No \\
\cite{pavlidis2000imaging} & Dual-band near infrared camera & No & No & No & No \\
\cite{zhang2011face} & Multi-spectral camera & No & No & No & No \\
\hline
\noalign{\smallskip}
\multicolumn{6}{c}{Multi-Modal} \\
\cite{frischholz2000biold} & Video and sound & No & No & Yes & No \\
\cite{chetty2006multi} & Video and sound & No & No & Yes & No \\
\hline
\noalign{\smallskip}
\multicolumn{6}{c}{Video analysis} \\
\cite{pan2007eyeblink} & Video & No & No & Yes & No \\
\cite{kollreider2005evaluating} & Video & No & No & Yes & No \\
\cite{bao2009liveness} & Video & No & No & Yes & No \\
\cite{wang2013face} & Video & No & No & Yes & No \\
\hline
\noalign{\smallskip}
\multicolumn{6}{c}{Two images under different lighting} \\
\cite{chan2018face} & Two images (flash / no-flash) & No & No & No & Yes \\
{\bf{ours}}     & Two images (diff. lighting conditions) & No & No & No & Yes\\
\hline
\noalign{\smallskip}
\multicolumn{6}{c}{Single image} \\
\cite{li2004live} & Single Photo & Yes & Yes & Yes & Yes \\
\cite{tan2010face} & Single Photo & Yes & Yes & Yes & Yes \\
\cite{peixoto2011face} & Single Photo & Yes & Yes & Yes & Yes \\
\cite{yeh2017face} & A camera with a shallow depth of field & Yes & Yes & Yes & Yes \\
\hline
\end{tabular}
\end{center}
\end{table*}
\setlength{\tabcolsep}{1.4pt}

\section{Conclusions}
In this work we introduced a new approach, and exemplified it with different methods, to overcome spoofing attacks and make face recognition technology more robust and reliable. The basic idea is to consider implicit 3D information, computed without the need to do a full 3D reconstruction. In particular, we focused on methods that could potentially be implemented in commercial devices such a personal computers or smartphones. We presented two different sets of features that can be practically applied due to their low hardware and computational requirements. Effective and discriminative features were engineered by exploiting the underling physical properties of 3D facial surfaces. We showed that from a pair of input images taken under different light conditions, implicit three dimensional features of the face can be captured. We compared these features with an approximated computation of the disparity obtained from an uncalibrated stereo pair of cameras. The proposed techniques were validated over different acquisition conditions, and we showed that the method generalizes across subjects, cameras, and scenes. Finally, we illustrated how particular input responses can be learned for different illumination conditions. This is a very interesting property from a practical point of view as it opens the possibility, for example, to create a more sophisticated authentication framework when multiple flashes could be arbitrary activated. In addition, validation experiments showed that the proposed features are invariant to the input images resolution and provide a more reliable facial representation than the existing alternatives.  

\section*{Acknowledgments}
Work partially supported by CSIC, NSF, NGA, ONR, and ARO.



\end{document}